\documentclass[letterpaper, 10 pt, conference]{ieeeconf}

\IEEEoverridecommandlockouts
\usepackage[caption=false, font=footnotesize]{subfig}
\usepackage{xcolor}
\usepackage{epsfig} %
\usepackage{graphics} %
\usepackage{times} %
\usepackage{amsmath} %
\usepackage{booktabs, multirow}
\usepackage{soul}%
\usepackage{changepage,threeparttable} %
\usepackage{amssymb}  %
\usepackage{subfig}
\usepackage{graphicx}
\usepackage{lipsum}  
\usepackage{amssymb}
\usepackage{array}
\usepackage{flushend}
\usepackage{cite}
\usepackage{hyperref}
\usepackage{float}
\usepackage{microtype}
\usepackage{xargs}
\usepackage{booktabs}
\usepackage{bm}
\usepackage{dsfont}

\usepackage[capitalise]{cleveref}
\usepackage{float}

\newcommand{\coloredline}[2]{\textcolor[HTML]{#1}{\raisebox{0.5ex}{\rule{#2}{2pt}}}}
\definecolor{myred}{HTML}{ff3030}

\title{\LARGE \bf 
\textcolor{myred}{RMMI}: \textcolor{myred}{R}eactive \textcolor{myred}{M}obile \textcolor{myred}{M}anipulation using an \textcolor{myred}{I}mplicit Neural Map}

\author{Nicolás Marticorena, Tobias Fischer, Jesse Haviland, Niko Suenderhauf %
\thanks{The authors are with the QUT Centre For Robotics, School of Electrical Engineering and Robotics at the Queensland University of Technology, Brisbane, QLD 4000, Australia. We acknowledge the ongoing support by the QUT Centre for Robotics and funding from ARC DECRA Fellowship DE240100149 to TF. Email: {\tt\small nicolas.marticorena@hdr.qut.edu.au}}}

\usepackage{amsmath}
\usepackage{amssymb}
\usepackage{accents}
\usepackage{bm}
\usepackage{xifthen}
\usepackage{color}
\usepackage{fancyvrb}
\usepackage{graphicx,scalerel}

\newcommand{\ba}{\begin{eqnarray}}
\newcommand{\ea}{\end{eqnarray}}

\newcommand{\presup}[1]{\,{}^{\scriptscriptstyle #1}\!}

\newcommand{\pose}[1][ZZZZ]{\ifthenelse{\equal{#1}{ZZZZ}}{}{\presup{#1}}{\mathbf{\xi}}}
\newcommand{\estpose}[1][ZZZZ]{\ifthenelse{\equal{#1}{ZZZZ}}{}{\presup{#1}}{\mathbf{\hat{\xi}}}}
\newcommand{\hpose}[1][ZZZZ]{\ifthenelse{\equal{#1}{ZZZZ}}{}{\presup{#1}}{\hat{\mathbf{\xi}}}}
\newcommand{\posedot}[1][ZZZZ]{\ifthenelse{\equal{#1}{ZZZZ}}{}{\presup{#1}}{\mathbf{\nu}}}

\newcommand{\q}[1][ZZZZ]{\ifthenelse{\equal{#1}{ZZZZ}}{}{\presup{#1}}{\mathring{q}}}

\DeclareMathAlphabet{\mathitbf}{OML}{cmm}{b}{it}
\newcommand{\twist}[2][ZZZZ]{\ifthenelse{\equal{#1}{ZZZZ}}{}{\presup{#1}}{\mathcal{S}}}
\renewcommand{\vec}[2][ZZZZ]{\ifthenelse{\equal{#1}{ZZZZ}}{}{\presup{#1}}{\mathitbf{#2}}}

\newcommand{\hvec}[2][ZZZZ]{\ifthenelse{\equal{#1}{ZZZZ}}{}{\presup{#1}}{\tilde{\vec{#2}}}}
\newcommand{\obvec}[2][ZZZZ]{\ifthenelse{\equal{#1}{ZZZZ}}{}{\presup{#1}}\rlap{${\overbridge{\phantom{$\vec{#2}$}}}$}\vec{#2}}
\newcommand{\evec}[2][ZZZZ]{\ifthenelse{\equal{#1}{ZZZZ}}{}{\presup{#1}}{\hat{\vec{#2}}}}
\newcommand{\bvec}[2][ZZZZ]{\ifthenelse{\equal{#1}{ZZZZ}}{}{\presup{#1}}{\bar{\vec{#2}}}}

\newcommand{\dvec}[2][ZZZZ]{\ifthenelse{\equal{#1}{ZZZZ}}{}{\presup{#1}}{\dot{\vec{#2}}}}
\newcommand{\ddvec}[2][ZZZZ]{\ifthenelse{\equal{#1}{ZZZZ}}{}{\presup{#1}}{\ddot{\vec{#2}}}}

\newcommand{\mat}[2][ZZZZ]{\ifthenelse{\equal{#1}{ZZZZ}}{}{\presup{#1}\,}{{\boldsymbol #2}}}
\newcommand{\dmat}[2][ZZZZ]{\ifthenelse{\equal{#1}{ZZZZ}}{}{\presup{#1}\,}{{\dot{\boldsymbol #2}}}}
\newcommand{\emat}[2][ZZZZ]{\ifthenelse{\equal{#1}{ZZZZ}}{}{\presup{#1}\,}{\hat{\boldsymbol#2}}}
\newcommand{\matfn}[3][ZZZZ]{\ifthenelse{\equal{#1}{ZZZZ}}{}{\presup{#1}}{{\mat{#2}}\left(#3\right)}}
\newcommand{\Rt}[2][ZZZZ]{\ifthenelse{\equal{#1}{ZZZZ}}{}{\presup{#1}}{{\bf R}\left(#2\right)}}

\newcommand{\point}[2][ZZZZ]{\ifthenelse{\equal{#1}{ZZZZ}}{}{\presup{#1}}{\mathbf{\mathrm{#2}}}}

\newfont{\School}{pncr}
\newfont{\eightTR}{pncr at 8pt}

\usepackage{color}
\usepackage{fancyvrb}
\fvset{formatcom=\color{blue},fontseries=c,fontfamily=courier,xleftmargin=4mm,commentchar=!}

\DefineVerbatimEnvironment{Code}{Verbatim}{formatcom=\color{blue},fontseries=c,fontfamily=courier,fontsize=\footnotesize,xleftmargin=4mm,commentchar=!}

\DefineVerbatimEnvironment{CodeSmall}{Verbatim}{formatcom=\color{blue},fontseries=c,fontfamily=courier,fontsize=\scriptsize,xleftmargin=1mm,commentchar=!}

\DefineVerbatimEnvironment{CodeNum}{Verbatim}{numbers=left,numbersep=4pt,formatcom=\color{blue},fontseries=c,fontfamily=courier,fontsize=\footnotesize,xleftmargin=4mm}

\newcommand{\model}[1]{\index{code}{#1@\textit{#1}}\ifthenelse{\boolean{draft}}{{\color{green}\Verb+#1+}}{\Verb+#1+}}
\newcommand{\block}[1]{\ifthenelse{\boolean{draft}}{{\color{green}\Verb+#1+}}{\textsf{#1}}}

\newcommand{\func}[2][ZZZZ]{\ifthenelse{\equal{#1}{ZZZZ}}{\index{code}{#2}}{\index{code}{#1}}\ifthenelse{\boolean{draft}}{{\color{green}\Verb+#2+}}{\Verb+#2+}}

\newcommand{\methodb}[2]{\index{code}{#1@\textbf{#1}!.#2}\ifthenelse{\boolean{draft}}{{\color{magenta}\Verb+#1.#2+}}{\Verb+#1.#2+}}
\newcommand{\method}[2]{\index{code}{#1@\textbf{#1}!.#2}\ifthenelse{\boolean{draft}}{{\color{magenta}\Verb+#2+}}{\Verb+#2+}}
\newcommand{\class}[1]{\index{code}{#1@\textbf{#1}}\ifthenelse{\boolean{draft}}{{\color{cyan}\Verb+#1+}}{\Verb+#1+}}
\newcommand{\property}[1]{\index{property}{#1}\ifthenelse{\boolean{draft}}{{\color{cyan}\Verb+#1+}}{\Verb+#1+}}

\usepackage{eso-pic}
\usepackage{url}
\AddToShipoutPicture*{%
     \AtTextUpperLeft{%
         \put(-3.5,10){
           \begin{minipage}{\textwidth}
              \scriptsize
              \MakeUppercase{IEEE/RSJ International Conference on Intelligent Robots and Systems (IROS) 2025.\\ Preprint version; final version available at} \url{https://ieeexplore.ieee.org}
           \end{minipage}}%
     }%
}

\begin{document}
\maketitle

\begin{abstract}
Mobile manipulator robots operating in complex domestic and industrial environments must effectively coordinate their base and arm motions while avoiding obstacles.
While current reactive control methods gracefully achieve this coordination, they rely on simplified and idealised geometric representations of the environment to avoid collisions. This limits their performance in cluttered environments.
To address this problem, we introduce RMMI, a reactive control framework that leverages the ability of neural Signed Distance Fields (SDFs) to provide a continuous and differentiable representation of the environment's geometry. RMMI formulates a quadratic program that optimises jointly for robot base and arm motion, maximises the manipulability, and avoids collisions through a set of inequality constraints. These constraints are constructed by querying the SDF for the distance and direction to the closest obstacle for a large number of sampling points on the robot.
We evaluate RMMI both in simulation and in a set of real-world experiments. For reaching in cluttered environments, we observe a 25\% increase in success rate. For additional details, code, and experiment videos, please visit \url{https://rmmi.github.io/}.
\end{abstract}

\section{Introduction}

Recent advances in perception and control have significantly enhanced the capabilities of reactive controllers, allowing robots to adapt their actions dynamically based on real-time sensor data. These controllers enable robust task execution by continuously adjusting the robot’s behaviour, facilitating complex operations such as visual servoing and object manipulation during motion~\cite{Haviland2022b, Burgess-Limerick2023a}.

Traditionally, reactive control has been valued for its speed and adaptability, though it was often considered less capable in handling complex environments compared to motion planning \cite{Zucker2013}, which, while more capable, is typically slower and computationally intensive. Recent works in motion planning~\cite{Thomason2023, Sundaralingam2023CuRoboGeneration}  have made significant strides in improving computational efficiency, allowing for faster planning in dynamic and cluttered environments. These advancements are narrowing the gap between reactive control and motion planning, making them more comparable in terms of speed and capability.

\begin{figure}[t]
        \centering
            \includegraphics[trim=0cm 0cm 0cm 0cm, clip, width=0.475\textwidth]{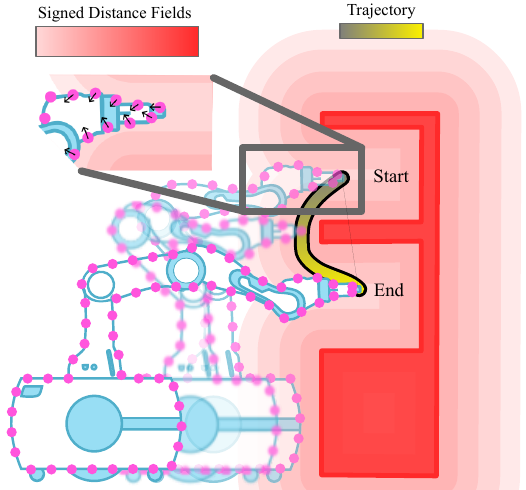}
        \caption{Illustration of our approach to reactive control in cluttered environments. The obstacles are represented within a Signed Distance Field (SDF), shown in red, where darker shades indicate closer proximity to the obstacle surfaces. The robot is represented as a set of points sampled from its outer geometry (depicted as magenta dots). Our controller computes the joint velocities required to follow a trajectory from the start to the end pose (highlighted in yellow), while simultaneously maximising manipulability and maintaining a safe distance from obstacles, all while staying within the constraints of joint limits and collision avoidance.}
        \label{fig:main_figure}
    \end{figure}

Despite advances in reactive controllers for mobile manipulation, they still face challenges in collision avoidance. Existing methods often rely on simplified scene models, assuming perfect knowledge of the environment using geometric primitives~\cite{Haviland2021}. Others extend 2D laser scan data vertically, which restricts the robot’s operation in confined spaces~\cite{Burgess-Limerick2023, Heins2021}. Some approaches incorporate sensor data to build scene models, but the discrete nature of these representations leads to approximations that limit the precision of collision detection~\cite{Pankert2020PerceptiveManipulation, Chiu2022, Mittal2022a}.

To address these limitations, we propose integrating a neural Signed Distance Field (SDF) as the underlying scene representation. Unlike traditional SDFs, neural SDFs provide a continuous representation of the environment, enabling more precise and efficient collision avoidance in cluttered settings.

The main contribution of this paper is RMMI, a reactive control framework for mobile manipulators that incorporates information from a neural SDF into the constraints of a Quadratic Program (QP) to enhance performance in complex environments.
Additionally, we introduce an active collision avoidance cost, which encourages the robot to maintain greater distances from obstacles, enhancing safety and further improving performance.
Through both simulation and real-world experiments, we demonstrate that our approach improves task success rates by at least 25\% over prior reactive controllers \cite{Haviland2021, Haviland2022b}.
Furthermore, it enables effective motion in cluttered and confined spaces using environment models constructed directly from onboard sensor data.

\section{Related Works}

\subsection{Collision Avoidance in Motion Generation}

Collision avoidance is a critical component in motion generation,
with seminal work introducing artificial potential fields to formulate collision avoidance as virtual repulsive fields~\cite{Khatib1985Real-timeRobots, Tulbure2020}.
Other approaches have incorporated collision avoidance within a Quadratic Programming (QP) framework, where it is treated as a set of hard inequality constraints~\cite{Haviland2021, Heins2021}.

While effective, these approaches often rely on simplified collision primitives or precise object models, limiting their applicability in unstructured environments~\cite{Khatib1985Real-timeRobots, Haviland2021}.
To address these limitations, some researchers have explored sensor-based approaches, using data from 2D laser scans on mobile manipulators~\cite{Heins2021, Burgess-Limerick2023} or depth maps for static manipulators~\cite{Vasilopoulos2023, Tulbure2020}. However, these approaches face challenges: 2D laser scans restrict the space in which a robot arm can operate, while depth maps are less practical for mobile manipulators due to limited viewing angles from robot-mounted cameras.

Our work advances these approaches by integrating a 3D representation that efficiently models collisions within a holistic reactive controller, eliminating the need for predefined geometric models and overcoming some of the limitations of purely sensor-based methods.

\subsection{Traditional Representations for Collision Avoidance}

To \textit{plan} collision-free motions, the robot requires a collision-checking routine; this routine needs to be \textit{efficient} given the large number of configurations that are tested during motion planning~\cite{Pan2012} and \textit{fast} to avoid lowering the controller frequency~\cite{Tulbure2020}. 
This has led to significant research into developing representations that support efficient collision checking, particularly those derived directly from sensor information.

One widely used representation is the \textit{occupancy-map}, where a 3D scene is represented as a grid of voxels, each indicating the probability of occupancy. A notable implementation is OctoMap~\cite{Hornung2013}, which segments the environment into ``occupied" or ``free" voxels, facilitating collision checking in sampling-based motion planning~\cite{Arbanas2018DecentralizedTeams}.

\subsection{Signed Distance Fields for Collision Checking}
Another prevalent approach is the use of signed distance fields (SDF), which store the distance to the nearest surface. SDFs offer the advantage of direct distance information as well as gradient information~\cite{Oleynikova2016}, making them suitable for planning methods such as trajectory optimisation~\cite{Zucker2013}. 
SDFs of a scene can be obtained by combining analytical distance functions~\cite{idealSDFs}, parameterisation inside a neural network~\cite{Park2019}, or stored in a voxel map~\cite{Oleynikova2017, Pan2022Voxfield:Reconstruction}. 
While voxel-based SDF maps have become popular in motion planning~\cite{Zucker2013} and model predictive control~\cite{Pankert2020PerceptiveManipulation,  Chiu2022, Mittal2022a}, they suffer from discretisation issues, necessitating trilinear interpolation or local linear models to refine distance estimates~\cite{Pankert2020PerceptiveManipulation}.

In contrast, our approach uses a continuous SDF parameterised by a neural network~\cite{Ortiz2022, Vasilopoulos2023HIO-SDF:Fields, Bolte2023},  which eliminates the fixed resolution of voxels and provides more accurate distance queries. Leveraging modern neural network libraries like PyTorch, we can perform multiple queries in parallel on a GPU, allowing for more precise robot representations compared to voxel-based approaches. Other work focused on learning SDFs of individual objects in the context of robotics control~\cite{Liu2022}.

Complementing environment-based approaches, some researchers have explored representing the robot itself as an SDF by learning its geometry using neural networks~\cite{Koptev2023} or basis functions~\cite{Li2023}. The robot's SDF enables precise control by computing the distance between the robot and its environment using static RGB-D cameras~\cite{Vasilopoulos2023} or external tracking systems~\cite{Li2023, Koptev2023}.

\section{Background}
This section provides an overview of the core components that underpin our approach: the holistic mobile manipulator motion controller (\cref{HollisticControl}) and the use of signed distance fields (SDFs) for precise environmental representation (\cref{sec:sdf}).

\subsection{Holistic Mobile Manipulator Motion Controller}\label{HollisticControl}

Our approach builds on the task space reactive controller introduced by~\cite{Haviland2022b}, which achieves desired end-effector poses by simultaneously using the degrees of freedom of the arm and the mobile base. This controller ensures that joint limits are respected and that the arm remains manoeuvrable throughout the trajectory. 

At the core of this controller is a Quadratic Program (QP) that is designed to output the required joint velocities $\dvec{q} \in \mathbb{R}^n$, where~$n$ represents the degrees of freedom of the mobile manipulator, considering the robot base as one revolute followed by a one prismatic virtual joint.
The primary objective is to achieve a desired spatial velocity $\vec{v} \in \mathbb{R}^6$ of the end-effector in the base frame $^b \vec{v}_e^{*}$, where $e$ and $b$ represent the end-effector and base frame. 
The optimisation problem is formulated as a QP as follows:
\begin{align}
    \min_{\vec{x}} \frac{1}{2} \vec{x}^\top \mat{Q} \vec{x} &+ \vec{c}^\top \vec{x}\label{eq:qp_inf}\\
    \text{s.t} \text{  }\mathcal{J}\vec{x} &= \vec[b]{v}_e \label{eq:equality}\\
    \mat{A} \vec{x} &\leq \vec{b} \label{eq:ab}\\
    \chi^{-} &\leq \vec{x} \leq \chi^{+}\label{eq:limits}
\end{align}
Here the decision variable $\vec{x}$ is a combination of the joint velocities $\dvec{q}$ and a slack component $\delta \in \mathbb{R}^6$:
\begin{equation}
    \vec{x} = \begin{pmatrix}\dvec{q}\\ \vec{\delta}\end{pmatrix} 
\in \mathbb{R}^{(n + 6)}.
\end{equation}

The quadratic cost function defined in~\cref{eq:qp_inf} is designed to minimise joint velocities, where the linear cost component $\vec{c}$ is given by:
\begin{equation}\label{eq:qp_c}
    \vec{c} = \begin{pmatrix}
        \mat{J}_m +  \mat{J}_o \\ \mathbf{0}_{6}
    \end{pmatrix} \in \mathbb{R}^{(n+6)},
\end{equation}
with $\mat{J}_m \in \mathbb{R}^n$ representing the manipulability Jacobian~\cite{Haviland2022c} that maximises the manipulability index of the robot arm~\cite{Yoshikawa1985a}, which indicates how well conditioned the current arm configuration is. The term $\mat{J}_o \in \mathbb{R}^n$ comprises a linear component to optimise the base's relative orientation to the end-effector~\cite{Haviland2022b}.

The equality constraints in \cref{eq:equality} are implemented through an augmented manipulator Jacobian $\mathcal{J}$~\cite{Haviland2022c}:
\begin{equation}
    \mathcal{J} = \begin{pmatrix}
        \mat[b]{J}_e & \mat{1}_{6 \times 6}
        \end{pmatrix} \in \mathbb{R}^{6 \times (n+6)},
\end{equation}
where $\mat[b]{J}_e \in \mathbb{R}^{6 \times n}$ maps joint velocities to the end-effector's spatial velocity relative to the base frame.

Inequality constraints in \cref{eq:ab} enforce joint limits, and \cref{eq:limits} defines bounds on the decision variable.

\subsection{Signed Distance Fields}\label{sec:sdf}

Signed distance fields (SDF) are a well-studied representation of objects and environments that is widely used in robotics~\cite{Oleynikova2016}, computer vision~\cite{Newcombe2011}, and computer graphics~\cite{Takikawa2022AFunctions}. 
In the 3D case, an SDF is a function $f\colon\mathbb{R}^3 \rightarrow \mathbb{R}$ that returns the distance $d \in \mathbb{R}$ of a sample point $\vec{p} \in \mathbb{R}^3$ to the nearest surface. The sign of this distance indicates if the sampled point $\vec{p}$ is inside the surface (negative), outside (positive), or on the surface (zero distance). Consequently, the surface $\mathcal{S}$ of the obstacle is represented by the zero-level set of the SDF:
\begin{equation}    
    \mathcal{S} = \left\{ \vec{p} \in \mathbb{R}^3 \mid f(\vec{p}) = 0 \right\}.
\end{equation}

Additionally, we can determine the direction to the closest obstacle by computing the gradient $\evec{g} \in \mathbb{R}^3$ of this function. This gradient vector points from the sampling point away from the surface, enabling the closest point on the surface $\vec{p}_{\text{near}}$ to be estimated as~\cite{Sharp2022}:
\begin{equation}
	\vec{p}_{\text{near}} = \vec{p} - d \cdot \evec{g} ,\text{ where } d = f(\vec{p}) \text{, and } \evec{g} = \nabla_{\vec{p}} f(\vec{p}).
\end{equation}
In our approach, SDFs are integral to accurately representing the robot's environment, allowing for precise collision avoidance within our holistic motion controller.

\section{Methodology}

Our controller is formulated as a QP optimisation problem, building on the principles discussed in \cref{HollisticControl}. 
In this context, we integrate collision avoidance as both individual constraints and a secondary optimisation objective that maximises the robot's overall distance from obstacles. The methodology comprises three main steps: (1) representing the robot’s geometry and querying the SDF based on sampled points from its outer mesh (\cref{sec:sampled_points}), (2) formulating collision constraints to ensure safe navigation (\cref{sec:collision_constrains}), and (3) introducing an active collision avoidance cost within the optimisation to enhance safety margins (\cref{sec:active_collision}).

\subsection{Robot Representation}\label{sec:sampled_points}
To effectively compute distances to obstacles using an SDF, the robot must be represented as a set of discrete points sampled from its outer mesh. These points, derived from the robot’s URDF model, provide a detailed approximation of the robot’s geometry.

\begin{figure}[t]
    \centering
        \centering
        \includegraphics[width=\linewidth]{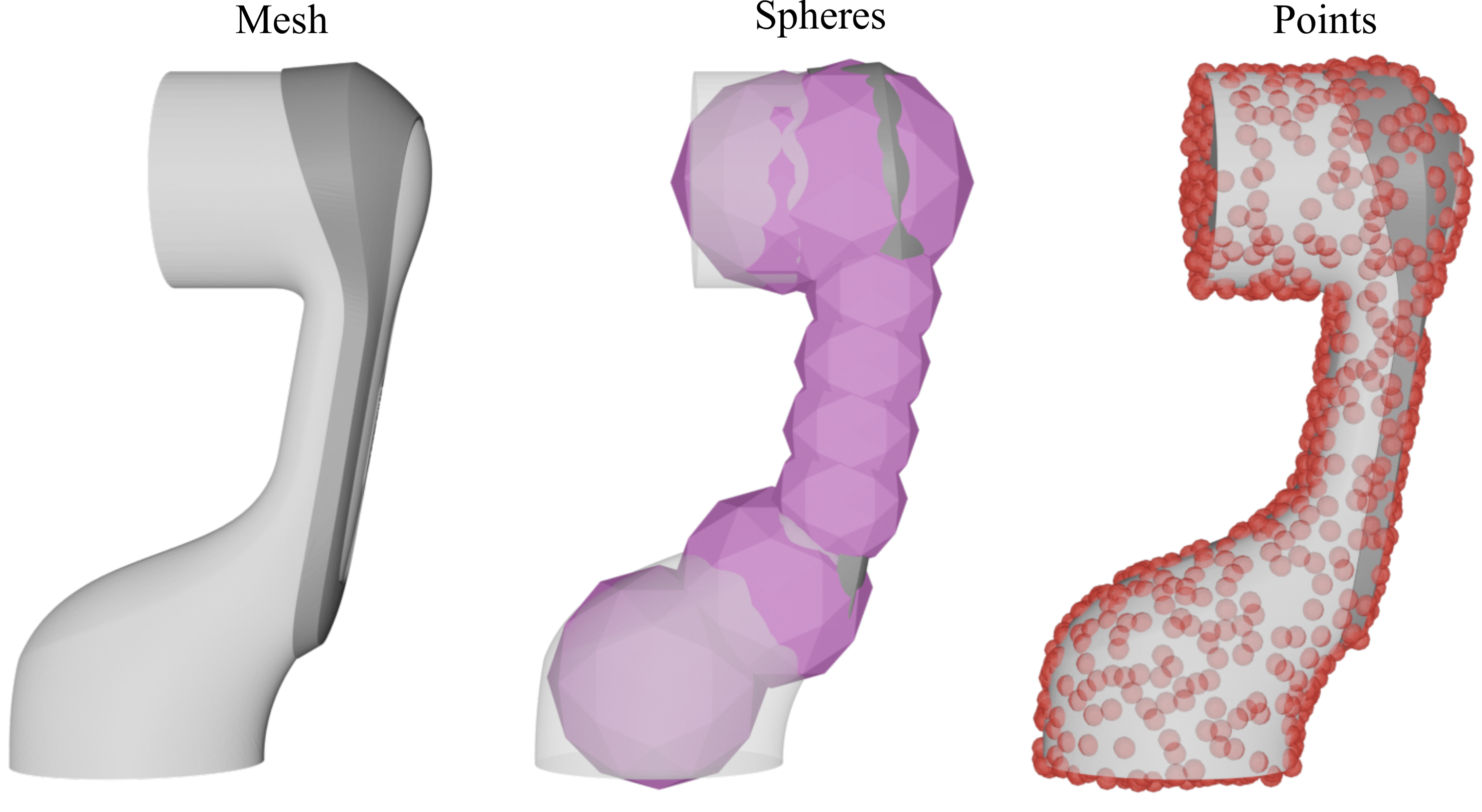}

    \caption{Comparison of different representations for one of the robot's links. The left image shows the original mesh geometry. The centre image illustrates a manual sphere-based representation, where the geometry is approximated by overlapping spheres, which can result in inflated and imprecise regions~\cite{Sundaralingam2023CuRoboGeneration}. The right image depicts our point-based representation, where points are sampled directly from the surface of the mesh, providing a more accurate and detailed representation of the robot's geometry.}
    \label{fig:sampling_points}
    \vspace{-0.2cm}
\end{figure}

In the first step, we group the various meshes corresponding to the robot's links according to the final joint in their respective kinematic chains. For each of these grouped meshes, we sample a set of points $\vec[k]{p}_j \in \mathbb{R}^{3}$ for the $j$-th point on the $k$-th link. The complete set of sampled points, $^{k}\mathcal{P}$ for the $k$-th link is represented as:
\begin{equation}    
    ^{k}\mathcal{P} = \left\{ ^{k}\vec{p}_{j} \mid j = 1, \cdots, M_{k} \right\},
\end{equation}
where $M_k$ is the total number of points sampled from the $k$-th link.

These points are then transformed into the world reference frame using the rigid body transformation $\mat[w]{T}_k \in SE(3)$ obtained through forward kinematics:
\begin{equation}
    ^{w}\mathcal{P} = \bigcup_{k} \left\{ \mat[w]{T}_k \cdot \phantom{}^{k}\vec{p}_j \mid \phantom{}^{k}\vec{p}_j \in \phantom{} ^{k}\mathcal{P} \right\}.
\end{equation}

This operation results in a set of $N$ points, where $N = \sum_{k}^{} M_k$ represents the total number of points sampled from the entire robot. An illustration comparing this point-based representation to a traditional sphere-based approach~\cite{Sundaralingam2023CuRoboGeneration} is shown in \cref{fig:sampling_points}, highlighting the precision of our method in capturing the robot’s geometry.

Selecting an appropriate value for $N$ involves balancing geometric fidelity and computational efficiency. With a smaller $N$, each sampled point must be expanded by a margin to approximate the robot's full geometry, which limits how closely it can approach obstacles. 
In contrast, increasing $N$ enhances geometric accuracy but raises computational costs. Our method scales linearly with the number of sampled points; however, this overhead is mitigated by parallelized batch queries for distance computations.

\subsection{Collision Constraints}\label{sec:collision_constrains}
With the robot represented as a set of sampled points and capable of querying the SDF for distance information, the next step is to incorporate these distances into the QP as hard constraints.
Following~\cite{Haviland2021}, we account for how joint movements affect the distances between the robot and obstacles. At every step $t$, the distance of the $j$-th sampled point to an obstacle is defined as:
\begin{equation}
    d_j(t) = f\big(\vec[w]{p}_{j}(t)\big).
\end{equation}
Moving forward, the subscript $j$ is omitted for ease of notation and compactness.

By taking the derivative of this distance with respect to time by applying the chain rule, we obtain
\begin{align}\label{eq:distance_derivate}
\dot{d}(t) =  \nabla_{\vec{p}} f(\vec{p}) \cdot \dvec[w]{p}(t) =  \evec{g} \cdot \dvec[w]{p}(t).
\end{align}
Here, $\dvec[w]{p}(t)$ represents the translational velocity of the point,  which can be expressed using the manipulator Jacobian $\mat{J}_{\nu}(\tilde{\vec{q}}) \in \mathbb{R}^{3\times k}$ considering each point as the end-effector~\cite{Haviland2022c}, where $\tilde{\vec{q}} = \left( q_0,q_1,...q_k \right) \in \mathbb{R}^k$ denotes the joints up to the $k$-th link.
Rewriting $\dvec[w]{p}(t)$ with respect to the joint's velocity results in:
\begin{align}\label{ps_der}
\dvec[w]{p}(t) = \mat{J}_{\nu}(\tilde{\vec{q}})\dot{\tilde{\vec{q}}}(t).
\end{align}
Substituting this into the derivative of the distance, we obtain:
\begin{align}
\evec{g} \cdot \mat{J}_\nu(\tilde{\vec{q}})\cdot\dot{\tilde{\vec{q}}} = \dot{d}(t),
\end{align}
which relates the change in distance to the joint velocities. Following \cite{Haviland2021}, we define the distance Jacobian $\mat{J}_{d_j}$ for each point: 
\begin{equation}    
    \mat{J}_{d_j}(\tilde{\vec{q}_j}) = \evec{g}_j \cdot \mat{J}_{\nu_{j}}(\tilde{\vec{q}_j}). 
\end{equation}
Where the subscript $j$ is now reintroduced for clarity.
Using a general velocity damper approach~\cite{Faverjon1987}, we impose the following inequality:
\begin{align}\label{eq:ine}
\mat{J}_{d_j} (\tilde{\vec{q}}) \dot{\tilde{\vec{q}}}(t) \leq \frac{ f(\vec[w]{p}_j) - d_s}{d_i - d_s},
\end{align}
where $d_i$ is the influence distance, i.e.~the distance at which the point is considered in the optimisation, and $d_s$ is the stopping distance, i.e.~the minimum allowable distance to an obstacle.

Stacking these inequalities for all points results in the following matrix-vector relation:
\begin{equation}   
    \begin{pmatrix}
        \mat{J}_{d_1}(\tilde{\vec{q}}_1) & \mathbf{0}_{1 \times 6 + n-k_1}\\ 
        \vdots & \vdots\\ 
        \mat{J}_{d_N}(\tilde{\vec{q}}_N) & \mathbf{0}_{1 \times 6 + n-k_N}\\  
    \end{pmatrix} \vec{x}(t) 
    \leq
    \begin{pmatrix} 
        \frac{f(\vec{p}_1) - d_s}{d_i - d_s} \\ 
        \vdots \\
        \frac{f(\vec{p}_N) - d_s}{d_i - d_s}
    \end{pmatrix}
\end{equation}
where the matrix on the left side is defined as $\mat{A}_c \in \mathbb{R}^{N \times (n+6)}$, and the right side vector as $\vec{b}_c \in \mathbb{R}^{N}$. Both can be integrated into the QP via \cref{eq:ab}.

\subsection{Active Collision Avoidance}\label{sec:active_collision}
Beyond the hard constraints, we introduce a collision avoidance cost into the objective function, promoting trajectories that maximise the robot's distance from obstacles.

To formulate this cost, we define a collision Jacobian $\mat{J}_c\in\mathbb{R}^{n}$ to represent how the distance between the robot and obstacles changes with respect to the joint velocities. This Jacobian is calculated as a weighted average of the individual distance Jacobians $\mat{J}_{d_{j}}$, with weights proportional to the distances:
\begin{equation}
    \mat{J}_c = \frac{\sum{\mat{J}_{d_{j}} (\vec{q}_j) w(\vec{p}_j)}}{\sum{w(\vec{p}_j)}},
\end{equation}
where the weight function $w(\vec{p}_j)$ is defined as:
\begin{equation}
     w(\vec{p}_j) = \frac{d_i - f(\vec{p}_{j})}{d_i - d_s}.
\end{equation}

\begin{table*}[t]
\centering
\caption{Results of the simulated reaching task, results are averaged across 1000 variations. Our approach achieves collision-free motions, whereas previous methods result in collisions due to the primitives not fully covering the robot's geometry. Additionally, the combination of both collision avoidance components in our method yields the highest success rates.}
\label{tab:main_table}
\begin{tabular}{@{}ccccccc@{}}
\toprule
Method &
  Collision model &
  Robot representation &
  \begin{tabular}[c]{@{}c@{}}Active Collision \\ Avoidance Cost \ref{sec:active_collision}\end{tabular} &
  \begin{tabular}[c]{@{}c@{}}Success \\ Rate (\%) $\uparrow$\end{tabular} &
  Collisions $\downarrow$ &
  $\left \langle |\vec{a}| \right \rangle (m/s^2) \downarrow$ \\ \midrule
Holistic \cite{Haviland2022b} &
  - &
  - &
  $\times$ &
  18.8 &
  81.2 &
  0.671 \\ \midrule
Baseline \cite{Haviland2021} + \cite{Haviland2022b} &
  Mesh-to-mesh &
  34 Primitives \cite{Corke2021} &
  $\times$ &
  50.2 &
  45.6 &
  0.740 \\
Baseline w active cost &
  Mesh-to-mesh &
  34 Primitives \cite{Corke2021} &
  $\checkmark$ &
  59.1 &
  35.3 &
  0.733 \\ \midrule
Ours w/o active cost &
  SDF &
  82 Spheres \cite{Sundaralingam2023CuRoboGeneration} &
  $\times$ &
  78.4 &
  0 &
  0.777 \\
Ours &
  SDF &
  82 Spheres \cite{Sundaralingam2023CuRoboGeneration} &
  $\checkmark$ &
  85.8 &
  0 &
  0.558 \\
Ours w/o active cost &
  SDF &
  2358 points &
  $\times$ &
  66.3 &
  0 &
  0.742 \\
Ours &
  SDF &
  2358 points &
  $\checkmark$ &
  79.5 &
  0 &
  \textbf{0.462} \\
Ours w/o active cost &
  SDF &
  9476 points &
  $\times$ &
  75.5 &
  0 &
  0.699 \\
Ours &
  SDF &
  9476 points &
  $\checkmark$ &
  \textbf{88.5} &
  0 &
  0.491 \\ \bottomrule
\end{tabular}
\end{table*}
However, $\mat{J}_c$ represents the rate of change of distance to the obstacle with respect to joint velocities. 
To incorporate a cost that depends on the robot's distance to the obstacle, we design a dynamic gain $\lambda_c$, inspired by the collision cost in CHOMP~\cite{Zucker2013}:
\begin{align}\label{eq:lamga_c}
    \lambda_c(t) = \frac{\lambda^{\max}_c}{\left(d_i - d_s \right)^2} \cdot \left( r_d(t) - d_i \right) ^2,
\end{align}
where $r_d$ is the distance of the robot defined as $\min_{j} f(\vec[w]{p}_j)$, and $\lambda^{\max}_c$ controls the steepness of the active collision gain.

Finally, we incorporate this collision cost into the overall optimisation:
\begin{equation}\label{eq:qp_c_with_col}
    \vec{c} = \begin{pmatrix}
        \mat{J}_m +  \mat{J}_o + \lambda_c \mat{J}_c \\ \mat{0}_{6}
    \end{pmatrix} \in \mathbb{R}^{(n+6)}.
\end{equation}
This modified optimisation encourages the robot to maintain a safer distance from obstacles while optimising for manipulability and orientation.

\section{Evaluation in Simulation}
To rigorously evaluate the effectiveness of our proposed method, we conducted extensive experiments in both simulated (this section) and real-world environments (\cref{sec:realworld}). The simulation experiments provide a controlled setting to test the performance under varying conditions, while the real-world experiments validate the practicality and robustness of our approach in real-time applications.

\subsection{Setup}

\subsubsection{Robot Platform}
For both the simulated and real-world experiments, we used Frankie\footnote{\url{ https://github.com/qcr/frankie_docs}}, a custom mobile manipulator platform consisting of a 7-degrees of freedom (DoF) Franka Panda arm mounted on a non-holonomic differential robot base, the Omron LD-60.

\subsubsection{Tasks}
The first set of experiments involved a reaching task, simulated using the Swift simulator and the Robotics Toolbox for Python~\cite{Corke2021} with a fixed step time of 0.05s (i.e, a control loop of 20Hz). The SDFs representing the obstacles were generated by combining analytic distance functions of primitive geometries~\cite{idealSDFs}. 
Using the ground truth geometry allows us to isolate the evaluation of our method from perception and SDF-generation errors, which are assessed in our real-world experiments in \Cref{sec:realworld}.

\begin{figure}[t]
    \centering
    \subfloat[Bookshelf]{%
        \includegraphics[width=0.22\textwidth, trim=15cm 2cm 15cm 10cm, clip]{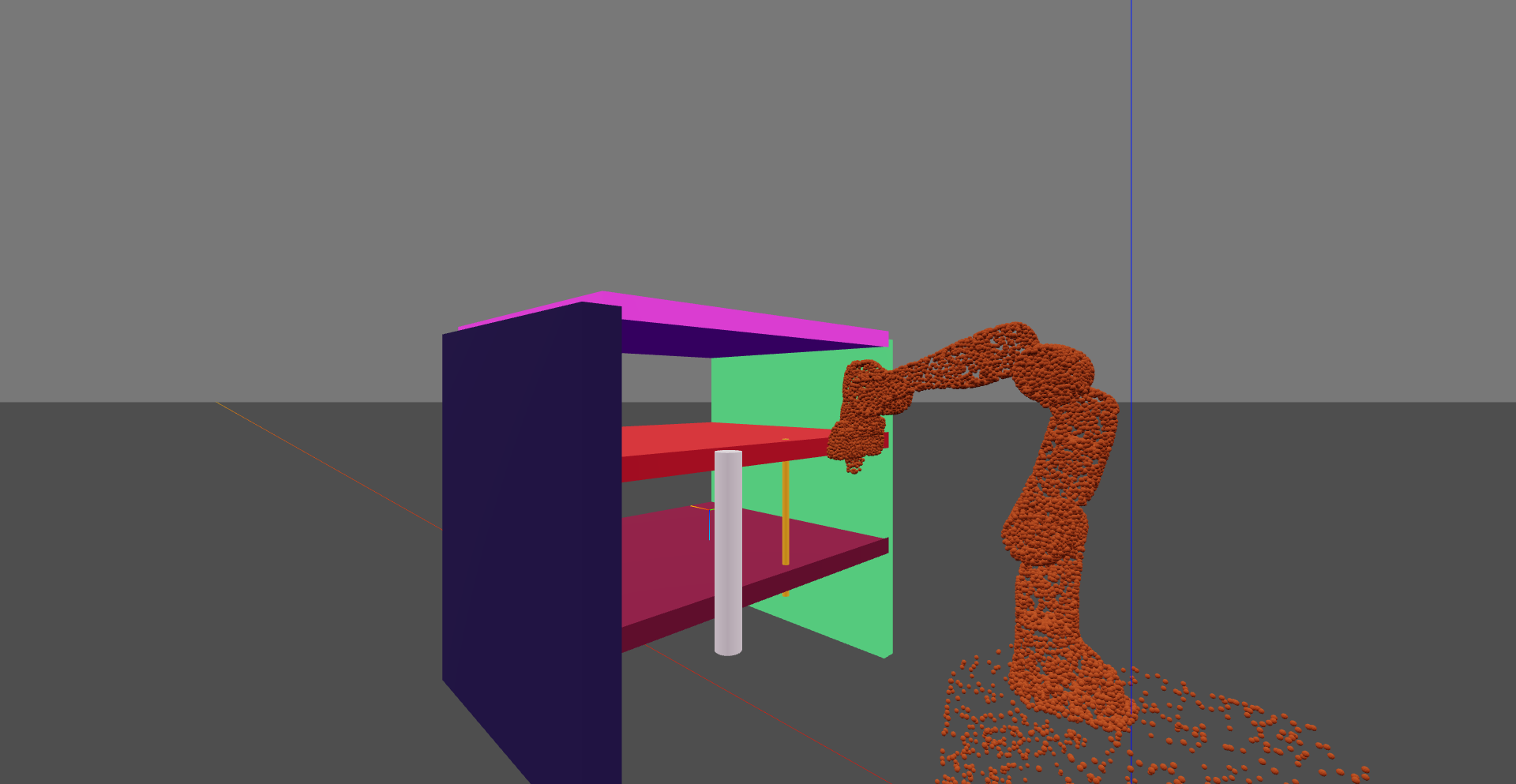}
        \label{subfig:Bookshelf}}
    \hfill
    \subfloat[Table\label{subfig:Table}]{%
        \includegraphics[width=0.22\textwidth, trim=15cm 2cm 15cm 10cm, clip]{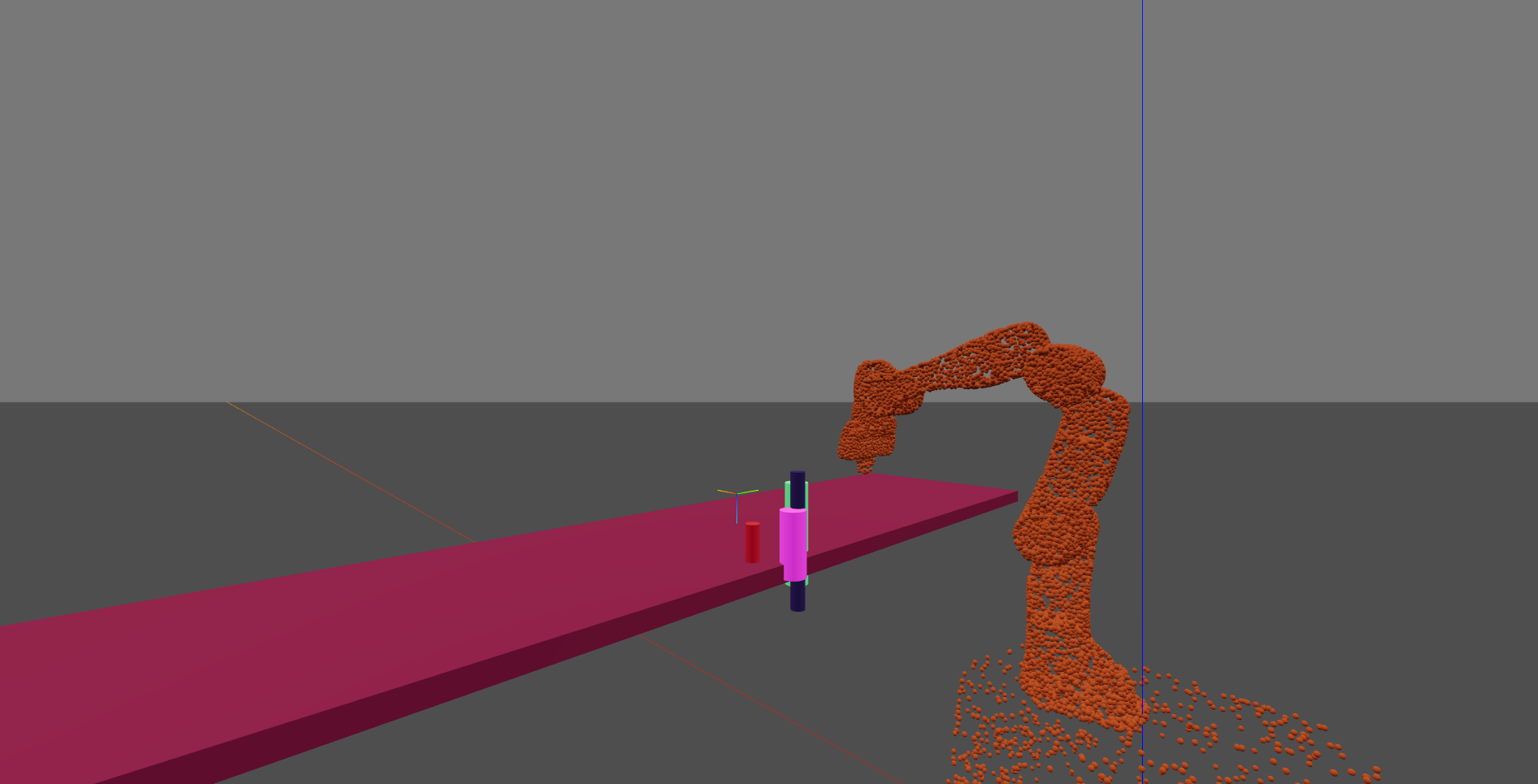}}
    \caption{Simulated environments in the Swift simulator: (a) Bookshelf and (b) Table. Each scene was varied across 500 iterations by randomly sampling the position and size of the obstacles, providing a diverse set of scenarios for evaluating the robot's performance in reaching tasks.}
    \label{fig:Simulated_envs}
    \vspace{-0.2cm}
    
\end{figure}

We designed two environments, denoted as Bookshelf and Table, as illustrated in \cref{fig:Simulated_envs}. In the Bookshelf environment (\cref{subfig:Bookshelf}), the heights of the top and bottom shelves, as well as the position and size of two cylinders placed on the front side of the bookshelf, were randomly varied. Similarly, in the Table environment (\cref{subfig:Table}), the height of the table and the position and size of four cylinders were randomly sampled.

We generated 500 different variations for each scenario, excluding any combinations where the target pose would result in a collision between the end-effector and obstacles.

Our proposed benchmark is available at \url{https://github.com/nmarticorena/frankie_planner}, which includes an RRT planner to verify the feasibility of the scenes.

\subsubsection{Metrics}
The following metrics were used to evaluate and compare the different approaches:
\begin{itemize}
    \item \textbf{Success:} The success rate of the controller in generating a collision-free trajectory that reaches the desired 6-DoF pose with a margin of 2cm as in previous work~\cite{Cohen2012}.

    \item \textbf{Collisions:} The rate of robot collisions with the environment, used to distinguish task failures caused by local minima from those caused by collisions.

    \item \textbf{Gracefulness:} Following previous work~\cite{Burgess-Limerick2023a} we measure the average absolute end-effector acceleration as a proxy for the smoothness of motion~\cite{gracefull}. We only consider successful motions to exclude cases where the robot stalled due to local minima.
\end{itemize}

\subsubsection{Baselines}
    We compare our approach against the holistic mobile manipulation controller presented in~\cite{Haviland2022b} and a modified version incorporating the collision avoidance component of NEO~\cite{Haviland2021}. This provided a baseline for comparing the use of an implicit SDF versus mesh-to-mesh collision queries using primitives.
Additionally, we tested our controller with two sets of sampled points and a modified version of the manually placed spheres approach from~\cite{Sundaralingam2023CuRoboGeneration} (modified to include the Omron LD-60 Base) to represent the robot. %
     Finally, because our active collision avoidance is built by combining multiple inequality constraints, we included an experiment using the baseline~\cite{Haviland2022b} that incorporates our active collision cost.

\subsubsection{Hardware specification}
All experiments were conducted on a machine with an Intel i7-12700K CPU, an NVIDIA RTX 3090 GPU, and 32GB RAM.

\subsection{Results}
The results of the simulation experiments are presented in \cref{tab:main_table}. Our approach, which integrates collisions constraints and a secondary cost function, achieved the highest performance, avoiding collisions in all attempts. When comparing different robot representations, our approach showed strong performance. Using a point representation with 9476 points, we achieved a success rate of 88.5\% (with the remaining 11.5\% getting stuck in local minima), followed by the manually placed spheres approach with 85.8\%. When reducing the number of sampled points, the performance gradually reduces, with a 79.5\% success rate when using only 2358 points (see also \cref{fig:control_rate}).

In contrast, running the controller without any collision awareness failed in 81\% of the runs. Modelling both the robot and the scene as geometric primitives yielded a success rate near 60\%, with a collision rate of at least 35\% due to the primitives not fully capturing the robot's geometry.  

\subsection{Ablation Studies}
To further investigate the robustness of our method, we conducted ablation studies focusing on the precision of point sampling and the combination of the different collision avoidance components in the optimisation.
    
    \subsubsection{Precision}\label{sec:precision}
    \begin{figure}[t]
        \centering
        \includegraphics[width = 0.85\linewidth]{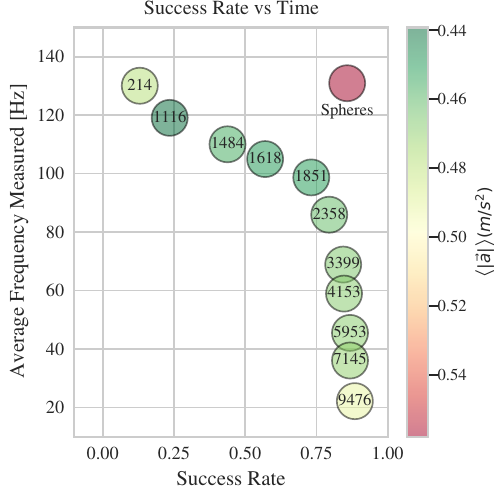}
            \label{subfig:rtx3090}
	    \caption{Comparison of point representations with varying number of sample points, where a faster control loop frequency is achieved at the cost of reduced performance in our benchmark tasks. The figure shows that as the number of sampled points decreases, the success rate drops, but the control loop operates at a higher frequency. Additionally, our approach using points instead of spheres results in smoother motions, as indicated by the lower average end-effector acceleration.}
        \label{fig:control_rate}
    \end{figure}

    We evaluated our approach using different levels of precision in the robot representation, achieved by varying the total number of sampled points. As shown in \cref{fig:control_rate}, reducing the number of sampled points increased the controller frequency at the cost of performance. The reduced precision required a higher stopping distance to prevent interpenetration, resulting in less information being available to the optimisation problem. For our real robot experiments, we selected the 2358 point representation, which offered the fastest controller frequency without reducing the success rate below 75\%.    

    \subsubsection{Effects of Active Collision}\label{sec:active}

\begin{figure}[t]
    \centering
(\coloredline{4d73b0}{0.2cm}) 9476 Points (\coloredline{c44e52} {0.2cm}) 2358 Points (\coloredline{54a768}{0.2cm}) Spheres

\includegraphics[width=0.9\linewidth]{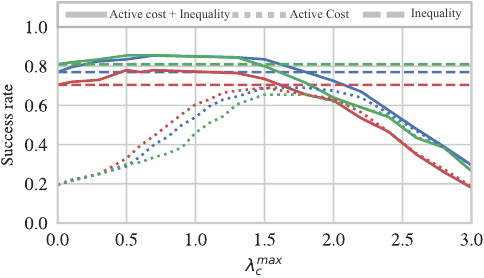}
\vspace{-0.2cm}
    \caption{Impact of varying the maximum gain $\lambda_c^{max}$ on the success rate, averaged across both simulated scenes. The results demonstrate that combining both inequality constraints and the active collision cost yields the highest performance across different robot representations.}
    
    \label{fig:active-collision}
\end{figure}

    We tested various combinations of our collision avoidance components, including a sweep on $\lambda_c^{\max}$ to control the gain of the active collision cost. These combinations were tested on an additional 100 variations of both simulated environments to avoid overlap between data used for hyperparameter tuning and method evaluation. The results are shown in \cref{fig:active-collision}, where the dashed lines represent the baseline without the active collision cost, the dotted lines show the performance with only active collision cost, and the solid lines depict our full proposed approach. The results indicate that using both the inequality constraints and the active collision cost improved the success rate across different sampled point variations and the sphere representation. However, $\lambda_c^{\max}\geq 1.5$ led to diminishing returns, as the robot began prioritising obstacle avoidance over moving towards the goal.
   
\begin{figure*}[t]
    \centering
    \centering
        \subfloat[Table]{
            \includegraphics[width=0.27\textwidth]{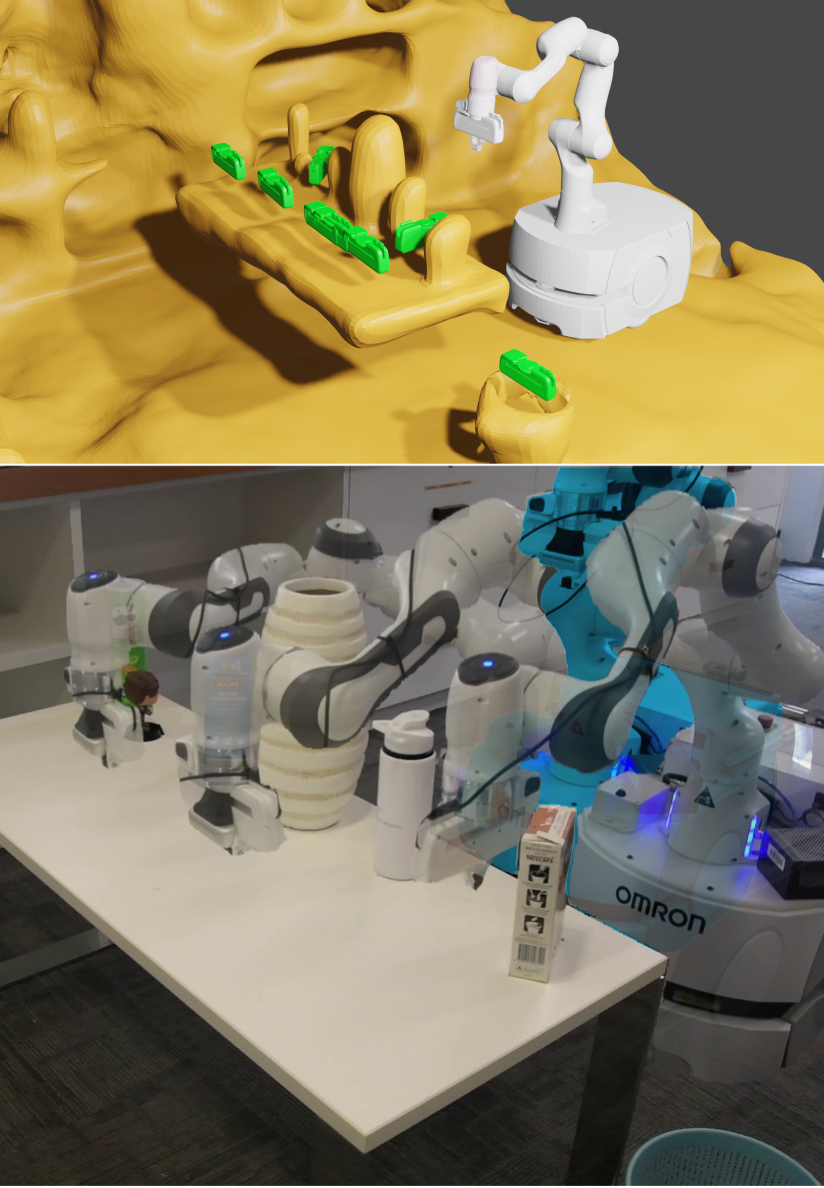}
            \label{subfig:RealTable}
        }
        \subfloat[Bookshelf]{
            \includegraphics[width=0.27\textwidth]{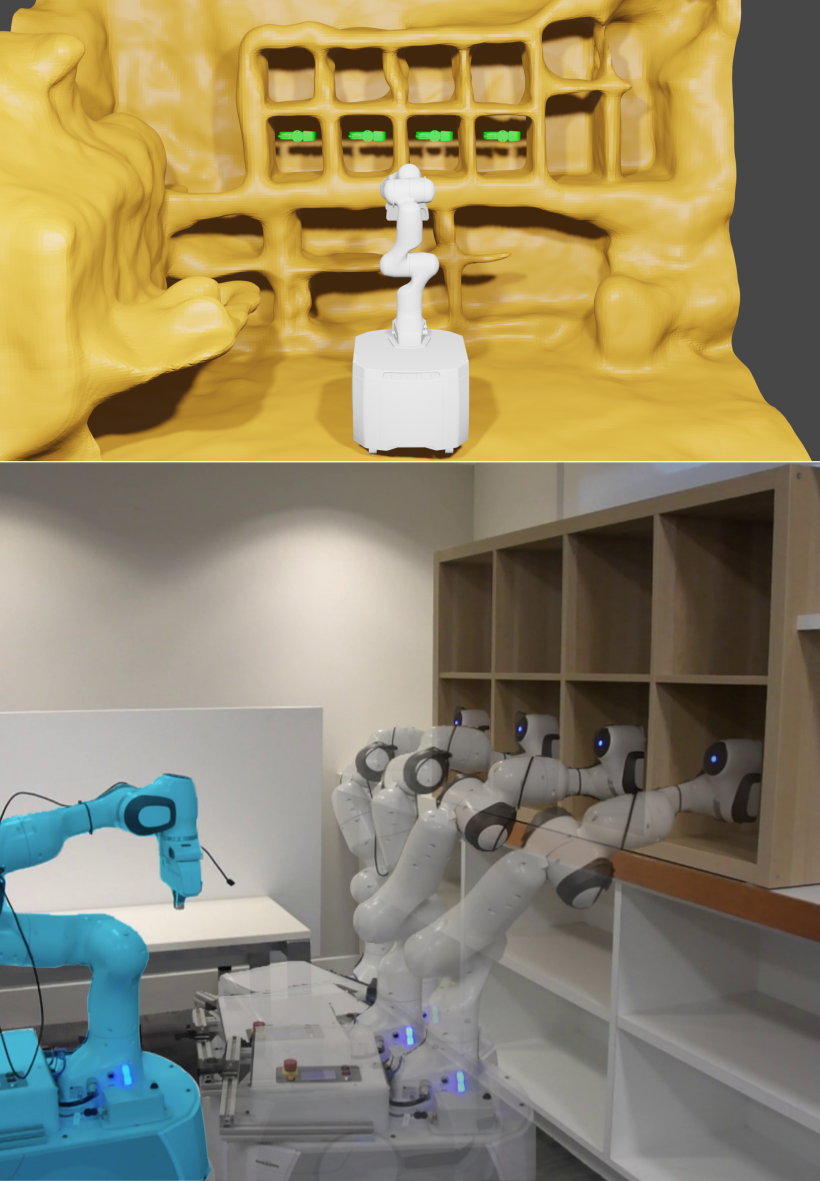}
            \label{subfig:RealBookshelf}
        }
        \subfloat[Cabinet]{
            \includegraphics[width=0.27\textwidth]{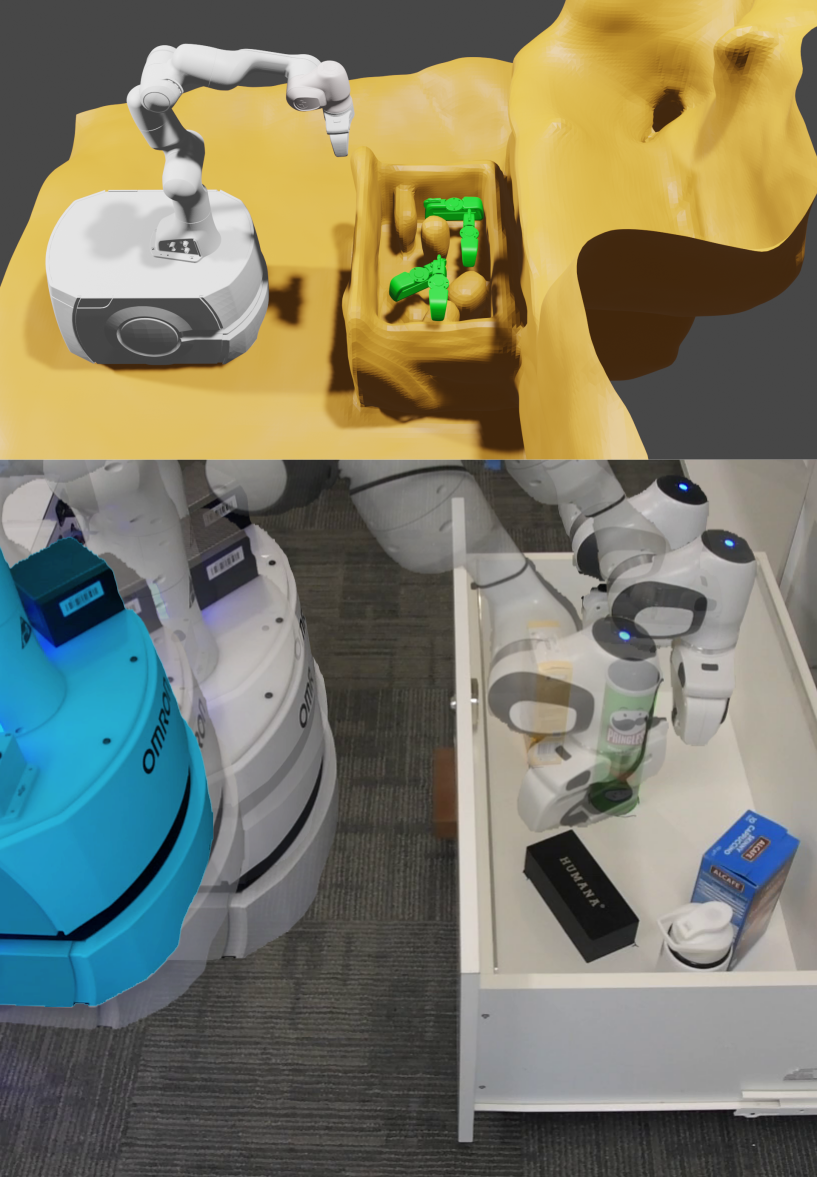}
            \label{subfig:RealCabinet}
        }
    \caption{(Top) Reconstructed meshes of the environments used for the real-world experiments: (a) Table, (b) Bookshelf, and (c) Cabinet. The green markers indicate the target 6-DoF end effector poses. (Bottom) Examples of the starting (blue) and final configurations of the robot during the experiments.}

        \label{fig:RealWorld}
        \vspace*{-0.2cm}
    \end{figure*}
\section{Real Robot Experiments}
\label{sec:realworld}

\begin{table}[t]
\centering
\caption{Real-world experiment results. Our approach successfully achieved collision-free motions across all scenarios. By combining both the inequality constraints and the active collision cost, we obtained the highest success rate.}
\label{tab:real_world}
\scriptsize
\begin{tabular}{@{}cccccc@{}}
\toprule
 &  & \begin{tabular}[c]{@{}c@{}}Collision \\ cost\end{tabular} & Success & Collisions & $\left \langle |a| \right \rangle (m/s^{2})$ \\ \midrule
\multirow{5}{*}{Bookshelf} & Holistic~\cite{Haviland2022b}     & $\times$     & 0/20  & 0/20  & \textbf{0.19} \\ \cmidrule(l){2-6} 
                           & \multirow{2}{*}{Points}  & $\times$     & 13/20 & 0/20  & 0.29 \\
                           &                          & $\checkmark$ & \textbf{20/20} & 0/20  & 0.39 \\ \cmidrule(l){2-6} 
                           & \multirow{2}{*}{Spheres} & $\times$     & 19/20 & 0/20  & 0.33 \\
                           &                          & $\checkmark$ & 19/20 & 0/20  & 0.36 \\ \midrule
\multirow{5}{*}{Cabinet}   & Holistic~\cite{Haviland2022b}     & $\times$     & 10/20 & 10/20 & 0.51 \\ \cmidrule(l){2-6} 
                           & \multirow{2}{*}{Points}  & $\times$     & \textbf{20/20} & 0/20  & 0.35 \\
                           &                          & $\checkmark$ & 19/20 & 0/20  & 0.31 \\ \cmidrule(l){2-6} 
                           & \multirow{2}{*}{Spheres} & $\times$     & 19/20 & 0/20  & 0.36 \\
                           &                          & $\checkmark$ & 19/20 & 0/20  & \textbf{0.24} \\ \midrule
\multirow{5}{*}{Table}     & Holistic~\cite{Haviland2022b}     & $\times$     & 14/40 & 26/40 & 0.51 \\ \cmidrule(l){2-6} 
                           & \multirow{2}{*}{Points}  & $\times$     & 28/40 & 0/20  & \textbf{0.28} \\
                           &                          & $\checkmark$ & 38/40 & 0/20  & 0.29 \\ \cmidrule(l){2-6} 
                           & \multirow{2}{*}{Spheres} & $\times$     & 30/40 & 0/20  & 0.31 \\
                           &                          & $\checkmark$ & \textbf{40/40} & 0/20  & 0.30 \\ \midrule
\midrule
\multirow{5}{*}{\textbf{Total}}     & Holistic~\cite{Haviland2022b}     & $\times$     & 24/80 & 36/80 & 0.40 \\ \cmidrule(l){2-6} 
                           & \multirow{2}{*}{Points}  & $\times$     & 61/80 & 0/80  & 0.31 \\
                           &                          & $\checkmark$ & 77/80 & 0/80  & 0.33 \\ \cmidrule(l){2-6} 
                           & \multirow{2}{*}{Spheres} & $\times$     & 68/80 & 0/80  & 0.34 \\
                           &                          & $\checkmark$ & \textbf{78/80} & 0/80  & \textbf{0.30} \\ \bottomrule
\end{tabular}

\end{table}

Following the promising results in simulation, we conducted experiments on a real robot to validate the practical effectiveness of our approach in complex, real-world environments.

\subsection{Robot Representation}
The robot representation used in the real-world experiments differs slightly from the simulation due to additional hardware considerations. These changes consider the geometry of the RealSense D435 mounted on the end effector and an inflated robot base to account for the Omron LD-60's internal safety margins.

\subsection{Tasks and Setup}
    We evaluated our method in three real-world scenarios, as shown in \Cref{fig:RealWorld}, with further details available in the supplementary video and on our website: \url{https://rmmi.github.io}. Each scene was reconstructed using iSDF~\cite{Ortiz2022} to generate the neural SDF from 30 pairs of posed depth images captured by the robot.

    During both the capture of the scene and the reaching task, the robot was localised in a 2D map using the in-built Omron-LD60 navigation capabilities.

    The experiments involved a reaching task with pre-defined target end-effector poses in three scenes. For the Cabinet and Bookshelf scenes, we tested four target poses, while the Table scene involved eight target poses. Each target pose was tested five times, resulting in a total of 80 trials per experiment. In each trial, the robot started with the arm in a home configuration, and the  base was positioned randomly in front of each scene.

    We compared our approach using both the manually placed spheres and our point-based representation, with and without our active collision avoidance cost. These methods were compared against a baseline controller without any collision awareness~\cite{Haviland2022b}.

\subsection{Results}
    Some examples of the starting configurations and target poses are shown in \cref{fig:RealWorld} and the attached video. The results across the three different scenes are presented in \cref{tab:real_world}. 
    Our approach, which incorporates both the inequality constraints and active collision cost, achieved the highest performance, with 96.3\% and 97.5\% for the point-based and sphere-based representations, respectively. Importantly, our method did not result in any collisions, with the remaining failures occurring only due to local minima.

    We observed the same trend regarding our active collision cost, observing an improvement in performance by at least 12.5\% for both point-based and sphere-based robot representations. Without collision avoidance, the robot collided in 45\% of the runs, with an additional 25\% of failures due to the robot becoming stuck when obstacles entered the safety zone of the base.

    Interestingly, the end-effector acceleration measurements revealed a 10\% decrease in acceleration when using the sphere-based representation compared to the point-based one. This contrasts with the simulation results, which showed higher acceleration for the sphere-based representation. We suspect this discrepancy may be due to the higher refresh rate of the real-world controller when using the sphere representation, whereas in simulation, a fixed rate was used across all experiments.

\section{Conclusions}
In this paper, we present a reactive controller capable of reliably reaching target poses while avoiding static obstacles, achieving a success rate of 96\% in real-world tasks. Leveraging a neural SDF, our approach efficiently integrates collision avoidance into the reactive controller while maintaining a fixed control rate. In addition, we introduce a simple yet effective active collision cost, which is robust to parameter variations and increases performance by at least 10\% in both simulation and real-world experiments.

Our approach effectively generates collision-free motions holistically, demonstrating robustness and efficiency. However, it is limited by the shortsightedness inherent in the local reactive controller paradigm, which makes it susceptible to local minima. This limitation could be mitigated by complementing our controller with a global planning strategy that updates at a lower frequency. Moreover, the parallel computation of distances enables our controller to operate in a receding horizon framework.

Our approach opens promising avenues for future research. One potential direction is to eliminate the assumption of a predefined camera path for reconstruction~\cite{Ortiz2022, Vasilopoulos2023HIO-SDF:Fields}. Integrating our controller with advances in the Neural Fields community~\cite{Xie2021} could allow for the autonomous generation of these representations using a robot. 
Additionally, recent work has explored novel distance fields that can be efficiently updated in the presence of dynamic obstacles~\cite{Ali2024}. While the core principles of our approach remain applicable, evaluating its performance in such dynamic scenarios is left for future work.

In conclusion, our reactive controller offers a robust and efficient solution for avoiding collisions with static obstacles. The approach provides a foundation for future research that could further enhance its capabilities and applicability in complex robotic tasks.

\bibliographystyle{IEEEtran} 
\bibliography{references, software}

\end{document}